\documentclass[11pt,letterpaper]{article}
\usepackage{authblk}
\usepackage{emnlp2017}
\usepackage{times}
\usepackage{latexsym}

\usepackage{flushend}

\emnlpfinalcopy

\usepackage{url}

\usepackage{booktabs}
\usepackage{graphicx}
\usepackage{amssymb}

\title{Fine-Grained Citation Span Detection for References in Wikipedia}

\author[1]{Besnik Fetahu}
\author[2]{Katja Markert}
\author[1]{Avishek Anand}

\affil[1]{
L3S Research Center, Leibniz University of Hannover\protect\\ 
Hannover, Germany\protect\\
{\tt \{fetahu, anand\}@L3S.de}
}
\affil[2]{
Institute of Computational Linguistics, Heidelberg University\protect\\
Heidelberg, Germany\protect\\
{\tt markert@cl.uni-heidelberg.de}
}

\date{}

\begin{document}
\maketitle

\begin{abstract}
\emph{Verifiability} is one of the core editing principles in
Wikipedia, editors being encouraged to provide citations for the added
content. For a Wikipedia article, determining the \emph{citation span}
of a citation, i.e. what content is covered by a citation, is important as it helps
decide for which content citations are still missing.

We are the first to address the problem of determining the \emph{citation span} in
Wikipedia articles. We approach this problem by classifying which
textual fragments in an article are covered by a
citation. We propose a sequence classification approach where for a
paragraph and a citation, we determine the citation span at a
fine-grained level.

We provide a thorough experimental evaluation and compare our approach against baselines adopted from the scientific domain, where we show improvement for all evaluation metrics.
\end{abstract}

\section{Introduction}\label{sec:introduction}

Citations uphold the crucial policy of \emph{verifiability} in Wikipedia. This policy requires Wikipedia contributors to support their additions with citations from authoritative external sources (web, news, journal etc.). 
In particular, it states that \emph{``articles should be based on reliable, third-party, published sources with a reputation for fact-checking and accuracy''\footnote{\url{https://en.wikipedia.org/wiki/Wikipedia:Identifying_reliable_sources}}}. 
Not only are citations essential in maintaining reliability, neutrality and authoritative assessment of content in such a collaboratively edited platform; but lack of citations are essential signals for core editors for unreliability checks.

\begin{figure}
	\centering
	\includegraphics[width=1.0\columnwidth]{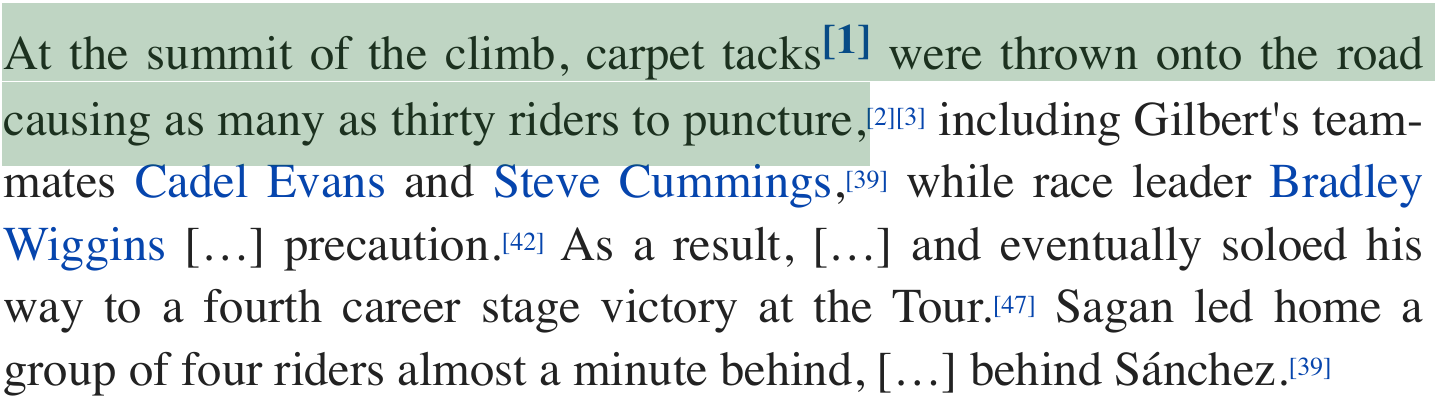}\vspace{-10pt}
	\caption{Sub-sentence level span for citation $^{\mathbf{[1]}}$ in a citing paragraph in a Wikipedia article.}
	\label{fig:span_example}
\end{figure}

\noindent However, there are two problems when it comes to citing
facts in Wikipedia.  First, there is a long tail of Wikipedia pages
where citations are missing and hence facts might be unverified.
Second, citations might have different span \emph{granularities},
i.e., the text encoding the fact(s), for which a citation is intended,
might span less than a sentence (see Figure~\ref{fig:span_example}) to
multiple sentences. We denote the different \emph{pieces of text}
which contain a citation marker as \emph{fact statements} or simply
\emph{statements}.  For example, Table~\ref{tbl:sentence_citation}
shows different \emph{statements} for several citations.  The
aim of this work is to automatically and accurately determine
\emph{citation spans} in order to improve
coverage~\cite{DBLP:conf/cikm/FetahuMA15,DBLP:conf/cikm/FetahuMNA16}
and to assist editors in  verifying citation quality at a
fine-grained level.

Earlier work on span determination is mostly concerned with scientific
texts~\cite{DBLP:journals/ipm/OConnor82,DBLP:journals/jip/KaplanTT16},
operates at sentence level and exploits explicit
authoring cues specific to scientific text. Although Wikipedia has
well formed text, it does not follow explicit scientific guidelines
for placing citations. Moreover, most statements can only be inferred
from the citation text.

In this work, we operate at a sub-sentence level, loosely referred to
as text fragments, and take a sequence prediction approach using a
\emph{linear chain CRF}~\cite{DBLP:conf/icml/LaffertyMP01}. We limit
our work to citations referring to \emph{web} and \emph{news} sources,
as they are accessible online and present the most prominent sources
in Wikipedia~\cite{DBLP:conf/websci/FetahuAA15}. By using recent work
on moving window language models~\cite{DBLP:conf/cikm/TanevaW13} and
 the structure of the paragraph
that includes a citation, we classify sequences of text fragments as
text that belong to a given citation. We are able to tackle all  citation
span cases as shown in Table~\ref{tbl:sentence_citation}.

\begin{table}[ht!]
\small
\begin{tabular}{p{1.5cm} p{5.5cm}}
\toprule
\texttt{sub sentence} & {Obama was born on August 4, $1961^{[c_1]}$, at Kapi'olani Maternity $\cdots$ Honolulu$^{[c_2]}$; he is the first $\cdots$ been born in Hawaii.$^{[c_3]}$}. \\[1.5ex]

\texttt{sentence} & {He was reelected to the Illinois Senate in 1998, $\cdots$ in 2002.$^{[c_1]}$} \\[1.5ex]

\texttt{multi sentence} & {On May 25, 2011, Obama $\cdots$ to address $\cdots$ UK Parliament in Westminster Hall, London. This was $\cdots$ Charles de Gaulle $\cdots$ and Pope Benedict XVI.$^{[c_1]}$} \\
\bottomrule
\end{tabular}
\caption{Varying degrees of citation span granularity in Wikipedia text.}
\label{tbl:sentence_citation}
\end{table}

\section{Problem Definition and Terminology}\label{sec:problem_definition}

In this section, we describe the terminology and define the problem of determining the \emph{citation span} in text in Wikipedia articles.

\paragraph{Terminology.} We consider Wikipedia articles $W=\{e_1,\ldots, e_n\}$ from a Wikipedia snapshot. We distinguish \emph{citations} to \emph{external references} in text and denote them with $\langle p_k, c_i\rangle$, where $c_i$ represents a citation which occurs in paragraph $p_k$ with positional index $k$ in an entity $e\in W$. We will refer to $p_k$ as the \emph{citing paragraph}. Furthermore, with \emph{citing sentence} we refer to the sentence in $s \in p_k$, which contains $c_i$.  Note that $p_k$ can have more than one citation as shown in Table~\ref{tbl:sentence_citation}.

\paragraph{Problem Definition.} The task of determining the \emph{citation span} for a citation $c$ and a paragraph $p$, respectively $\langle p, c\rangle$ (or simply $p_c$), is subject to the citing paragraph and the citation content. In particular, we refer with \emph{citation span} to the \emph{textual fragments} from $p$ which are covered by $c$. The fragments correspond to the sequence of \emph{sub-sentences} $\mathcal{S}(p)=\langle \delta_1^1, \delta_1^2,\ldots,\delta_1^k,\ldots,\delta_n^m\rangle$. We obtain the sequence of sub-sentences from $p$ by splitting the sentences into sub-sentences or text fragments based on the following punctuation delimiters ($\{,!;:?\}$). These delimitors do not always provide  a perfect semantic segmentation of sentences into facts. A more involved approach could be taken akin to work in text summarization, such as  Zhou and Hovy~\cite{zhou2006summarization} or \cite{DBLP:journals/tslp/NenkovaPM07} who  consider \emph{summary units} for a similar purpose.

Formally, we define the \emph{citation span} in
Equation~\ref{eq:cite_span} as the function of finding the subset
$\mathcal{S}'\subseteq\mathcal{S}$ where the fragments in
$\mathcal{S}'$ are covered by $c$.
\begin{equation}\label{eq:cite_span}
\varphi(p, c)\rightarrow \mathcal{S}'\subseteq\mathcal{S}, \;\; s.t.\;\; \delta \in \mathcal{S}' \wedge c \vdash \delta
\end{equation}
where $c \vdash \delta$ states that $\delta$ is covered in $c$.

\section{Related Work}\label{sec:related_work}

\textbf{Scientific Text.} One of the first attempts to determine the
citation span in text~\cite{DBLP:journals/ipm/OConnor82} was carried
out in the context of document retrieval. The citing statements from a
document were used as an index to retrieve the \emph{cited}
document. The citing statements are extracted based on heuristics
starting from the citing sentence and are expanded with sentences in a
window of +/-2 sentences, depending on them containing cue words like
\emph{`this', `these',$\ldots$ `above-mentioned'}. We consider the
approach in~\cite{DBLP:journals/ipm/OConnor82} as a baseline.

Kaplan et al.~\shortcite{DBLP:journals/jip/KaplanTT16} proposed the
task of determining the \emph{citation block} based on a set of
\emph{textual coherence} features (e.g. grammatical or lexical
coherence). The citation block \emph{starts} from the citing sentence,
with succeeding sentences classified (through SVMs or CRFs) according
to whether they belong to the block. Abu-Jbara and
Radev~\shortcite{DBLP:conf/naacl/Abu-JbaraR12} determine the citation
block by first segmenting the sentences and then classifying
individual words as being \emph{inside}/\emph{outside} the
citation. Finally, the segment is classified depending on the word
labels (majority of words being inside, at least one, or all of
them). This approach is not applicable in our case due to the fact
that words in Wikipedia text are not domain or genre-specific as one
expects in scientific text, and as such their classification does not
work.

\textbf{Citations in IR.} The importance of determining the citation
span has been acknowledged in the field of Information Retrieval
(IR). The focus is on building citation indexes~\cite{Garfield108} and
improving the retrieval of scientific
articles~\cite{DBLP:conf/cikm/RitchieRT08,Ritchie:2006:FBI:1629808.1629813}. Citing
sentences on a fixed window size are used to index documents and aid
the retrieval process.

\textbf{Summarization.} Citations have been successfully employed to
generate summaries of scientific
articles~\cite{DBLP:conf/coling/QazvinianR08,DBLP:journals/jasis/ElkissSFESR08}. In
all cases, citing statements are either extracted manually or via
heuristics such as extracting only citing
sentences. Similarly~\cite{DBLP:conf/ijcai/NanbaO99} expand the
summaries in addition to the citing sentence based on cue words
(e.g. \emph{`In this', `However'} etc.). The work
in~\cite{DBLP:conf/acl/QazvinianR10} goes one step beyond and
considers sentences which do not \emph{explicitly} cite another
article. The task is to assign a binary label to a sentence,
indicating whether it contains context for a cited paper. We use this
approach as one of our competitors. Again,  the premise is
that citations are marked explicitly and additional citing sentences
are found dependent on them.

\textbf{Comparison to our work.}  The language style and the
composition of citations in Wikipedia and in scientific text differ
significantly. Citations are \emph{explicit} in scientific text
(e.g. \emph{author names}) and are usually the first word in a
sentence~\cite{DBLP:conf/naacl/Abu-JbaraR12}. In Wikipedia, citations
are \emph{implicit} (see Table~\ref{tbl:sentence_citation}) and there
are no cue words in text which link to the provided citations.
Therefore, the proposed methodologies and features from the scientific
domain do not perform optimally in our case.

Both \cite{DBLP:conf/acl/QazvinianR10} and
\cite{DBLP:journals/ipm/OConnor82} work at the sentence level.  As, in
Wikipedia, citation span detection needs to be performed at the
sub-sentence level (see Table~\ref{tbl:sentence_citation}), their
  method introduces erroneous spans as we will show in our evaluation.

Related to our problem is the work on addressing quotation
attribution. Pareti et al.~\shortcite{DBLP:conf/emnlp/ParetiOKCK13}
propose an approach for \emph{direct} and \emph{indirect} quotation
attribution. The task is mostly based on lexical cues and specific
\emph{reporting verbs} that are the signal for the majority of direct
quotations. However, in the case of quotation attribution the task is
to find the \emph{source, cue}, and \emph{content} of the quotation,
whereas in our case, for a given citing paragraph and reference we
simply assess which text fragment is covered by the reference. We also
do normally not have access to specific lexical links between the citation
and its citation span.

\section{Citation Span Approach}\label{sec:approach}

We approach the problem of citation span detection in Wikipedia as a
\emph{sequence classification} problem. For a citation $c$ and a
citing paragraph $p$, we chunk the paragraph into textual fragments at
the \emph{sub-sentence} granularity, shown in
Equation~\ref{eq:cite_span}.

\begin{figure}[ht!]
\centering
\includegraphics[width=1.0\columnwidth]{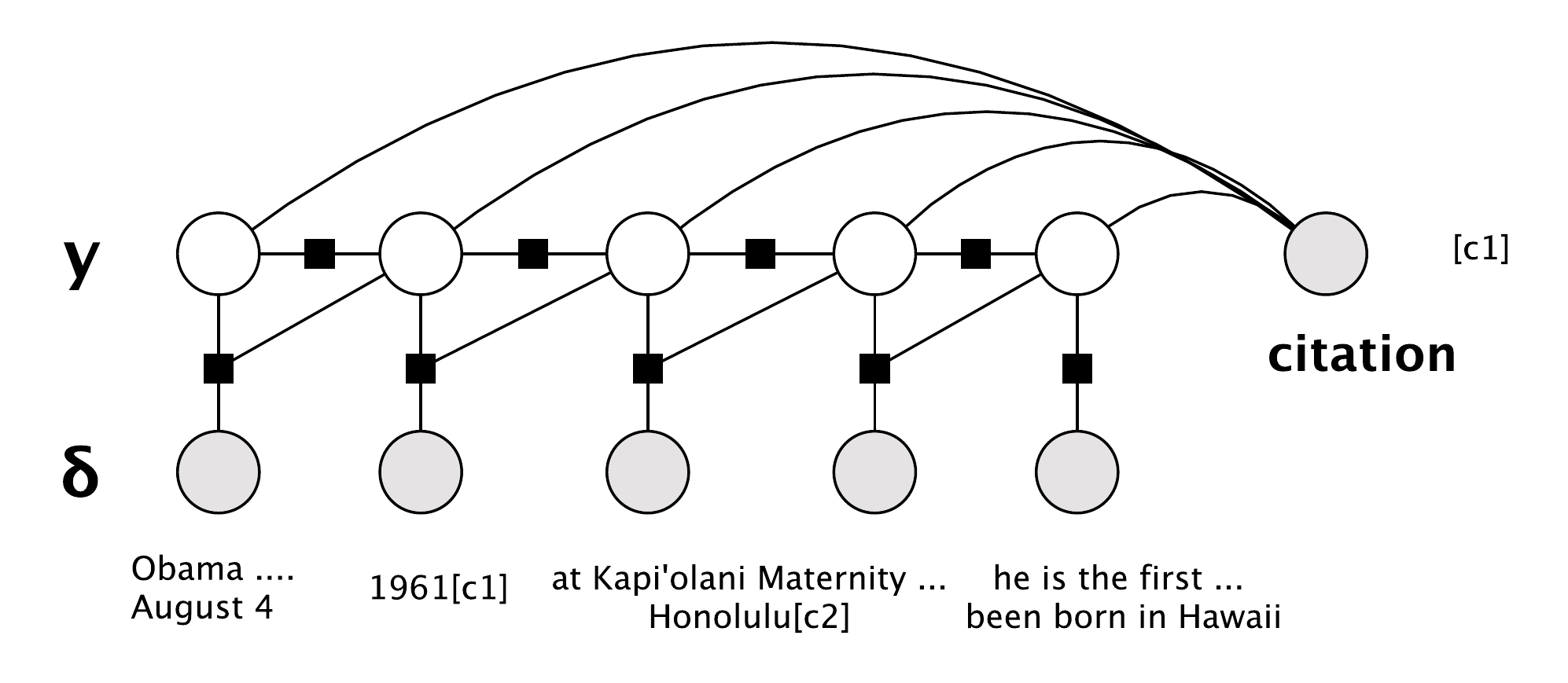}
\caption{Linear chain CRF representing the sequence of text fragments in a paragraph. In the factors we encode the fitness to the given citation.}
\label{fig:crf_chain}
\end{figure}

Figure~\ref{fig:crf_chain} shows an overview of the sequence
classification of textual fragments. We use a \emph{linear chain
  CRF}~\cite{DBLP:conf/icml/LaffertyMP01}, where for any fragment
$\delta$ we predict the label corresponding to a random variable
$\textbf{y}$ which is either \emph{`covered'} or
\emph{`not-covered'}. We opt for CRFs since we can encode global
dependencies between the text fragments and the actual citation, thus,
ensuring the coherence and accuracy of the predicted labels.

We now describe the features we compute for the factors
$\Psi(y_i,y_{i-1}, \delta_i)$ for a fragment $\delta_i$ w.r.t the
citation $c$. We determine the fitness of $\delta_i$ holding true or
being covered by $c$. We denote with $f_k$ the features for the
factors $\Psi_i(y_{i}, y_{i-1}, \delta_i)$ for sequence $\delta_i$ for
the linear-chain CRF in Figure~\ref{fig:crf_chain}.

\subsection{Structural Features}\label{subsec:struct_factors}

An important aspect to consider for citation span detection is the
structure of the citing paragraph, and correspondingly its
sentences. For a textual fragment $\delta$, we extract the following
structural features shown in Table~\ref{tbl:struct_factors}.

\begin{table}[ht!]
\centering\small
\begin{tabular}{l p{5cm}}
\toprule
factor & description\\
\midrule
$f_i^{c'}$ &  presence of other citations in $\delta_i$ where $c'\neq c$\\
$f^{\#s}$ & the number of sentences in $p$\\
$f_i^{|\delta_i|}$ & the length in terms of characters of the sub-sequence\\
$f_i^{s}$ &  check if $\delta_i$ is in the same sentence as the citation $c$\\
$f_i^{s\neq s'}$   & check if $\delta_i$ is in the same sentence as $\delta_{i-1}$\\
$f_i^{c}$ &  the distance of fragment $\delta_i$ to the fragment which contains citation $c$\\
\bottomrule
\end{tabular}
\caption{Structural features for a fragment $\delta_i$.}
\label{tbl:struct_factors}
\end{table}

From the features in Table~\ref{tbl:struct_factors}, we highlight
$f_i^{c}$ which specifies the distance of $\delta$ to the fragment
that cites $c$. The closer a fragment is to the citation the higher
the likelihood of it being covered in $c$. In Wikipedia, depending on
the citation and the paragraph length, the validity of a citation is
densely concentrated in its nearby sub-sentences (preceding and
succeeding).

Furthermore, the features $f^{\#s}$ and $f_i^{s}$ (the
number of sentences in $p$ together with the feature considering if $\delta$
is in the same sentence as $c$) are strong
indicators for accurate prediction of  the label of $\delta$. That is, it
is more likely for a fragment $\delta$ to be covered by the citation
if it appears in the same sentence or sentences nearby to the citation
marker.

However, as shown in Table~\ref{tbl:sentence_citation} there are three
main citation span groups, and as such relying only on the structure
of the citing paragraph does not yield optimal results. Hence, in the
next group we consider features that tie the individual fragments in
the citing paragraph with the citation as shown in
Figure~\ref{fig:crf_chain}.

\subsection{Citation Features}\label{subsec:cite_features}

A core indicator as to whether a fragment $\delta$ is covered by $c$
is based on the lexical similarity between $\delta$ and the content in
$c$. We gather such evidence by computing two  similarity
measures. We compute the features $f_i^{LM}$ and $f_i^{J}$ between
$\delta$ and paragraphs in the citation content $c$.

The first measure, $f_i^{LM}$, corresponds to a moving language window
proposed in \cite{DBLP:conf/cikm/TanevaW13}. In this case, for each
word in either a paragraph in the citation $c$ or the sequence
$\delta$, we associate a language model $M_{w_i}$ based on its context
$\phi(w_i)=\{w_{i-3}, w_{i-2}, w_{i-1}, w_i, w_{i+1}, w_{i+2},
w_{i+3}\}$ with a window of +/- 3 words. The parameters for the model
$M_{w_i}$ are estimated as in Equation~\ref{eq:lm_model} for all the
words in the context $\phi(w_i)$ and their frequencies denoted with
$tf$. With $M_\delta$ and $M_p$ we denote the overall models as
estimated in Equation~\ref{eq:lm_model} for the words in the
respective fragments.
\begin{equation}\label{eq:lm_model}\small
P(w|M_{w_i})= \frac{tf_{w,\phi(w_i)}}{\sum_{w'\in \phi(w_i)}tf_{w',\phi(w_i)}}
\end{equation}
Finally, we compute the similarity of each word in $w \in \delta$ against the language model of paragraph $p \in c$ in Equation~\ref{eq:lm_sim_feature}, which corresponds to the Kullback-Leibler divergence score.
\begin{equation}\label{eq:lm_sim_feature}\small
f_i^{LM} = \min\limits_{p \in c}\left[-\sum_{w \in \delta} P(w|M_{\delta}) \log{\frac{P(w|M_{\delta})}{P(w|M_p)}}\right]
\end{equation}

The intuition behind $f_i^{LM}$ is that for the fragments $\delta$ we
take into account the word similarity and the similarity in the
context they appear in w.r.t a paragraph in a citation. In this way, we
ensure that the similarity is not by chance but is supported by the
context in which the word appears. Finally, another advantage of this
model is that we localize the paragraphs in $c$ which provide evidence
for $\delta$.

As an additional feature we compute $f_i^{J}$ which corresponds to the
maximal \emph{jaccard} similarity between $\delta_i$ and paragraphs
$p\in c$.

Finally, as we will show in our experimental evaluation in
Section~\ref{sec:setup}, there is a high correlation between the
citation span length and the length of citation content in terms of
sentences. Hence, we add as an additional feature $f^c$ the number of
sentences in $c$.

\subsection{Discourse Features}\label{subsec:disc_factors}
Sentences and fragments within a sentence can be tied together by 
discourse relations. 
 We annotate  sentences with
explicit discourse relations based on an approach proposed in
\cite{DBLP:conf/acl/PitlerN09}, using discourse connectives as cues.
The explicit discourse relations belong to one of the
following: \emph{temporal, contingency, expansion, comparison}.

After extracting a discourse connective in a sentence, we determine by
its position to which fragment it belongs and mark the fragment
accordingly.  We denote with $f_i^{disc}$ the discourse feature for
the fragment $\delta_i$.\footnote{Note that, although discourse
  relations hold between at least two fragments or sentences, we only
  mark the individual fragment in which the connective occurs with the
  discourse relation type.}

\subsection{Temporal Features}\label{subsec:temp_factors}

An important aspect that we consider here is the temporal difference
between two consecutive fragments $\delta_i$ and $\delta_{i-1}$.  If
there exists a temporal date expression in $\delta_i$ and
$\delta_{i-1}$ and they point to different time-points, this presents
an indicator on the transitioning between the states $y_i$ and
$y_{i-1}$. That is, there is a higher likelihood of changing the state
in the sequence $\mathcal{S}$ for the labels $y_i$ and $y_{i-1}$.

We compute the temporal feature  $f_i^{\lambda(i,i-1)}$,indicating
the difference in \emph{days} between any two temporal expression
extracted from $\delta_i$ and $\delta_{i-1}$. We extract the temporal
expression through a set of hand-crafted regular expressions. We use
the following expressions: (1) \texttt{DD Month YYYY}, (2) \texttt{DD
  MM YYYY}, (3) \texttt{MM DD YY(YY)}, (4) \texttt{YYYY}, with
delimiters (whitespace, `-', `.').

\section{Experimental Setup}\label{sec:setup}

We now outline the experimental setup for evaluating the citation span approach and the competitors for this task. The data and the proposed approaches are made available at the paper URL\footnote{\url{http://l3s.de/~fetahu/emnlp17/}}.

\subsection{Dataset}\label{subsec:dataset}

\begin{figure}[ht!]
\centering
\includegraphics[width=0.8\linewidth]{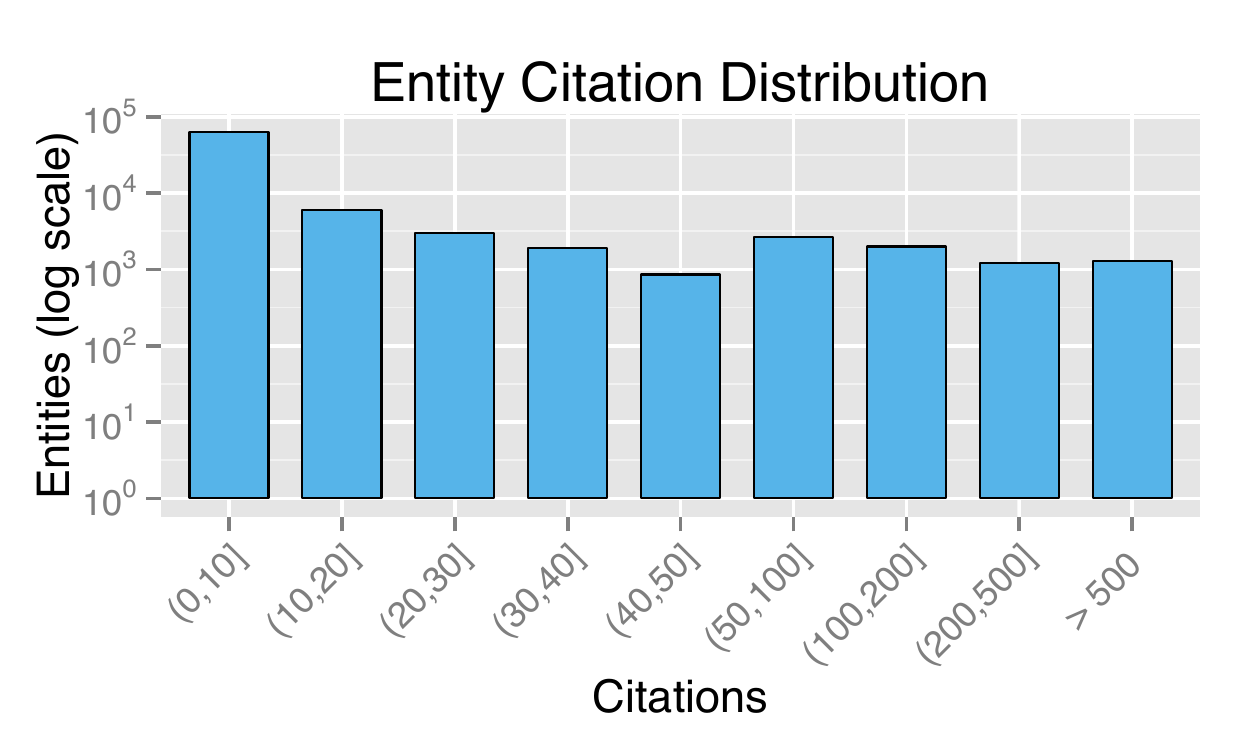}
\caption{Entity distribution based on the number of news citations.}
\label{fig:entity_news_cite_dist}
\end{figure}

We evaluate the citation span approaches on a random sample of
Wikipedia entities (snapshot of 20/11/2016). For the sampling process,
we first group entities based on the number of \emph{web}
or \emph{news} citations.\footnote{Wikipedia has an internal categorization of
  citations based on the reference they point to.}). We then  sample from
the specific groups. This is due to the inherent differences in
citation spans for entities with different numbers of citations. For
instance, entities with a high number of citations tend to have shorter
spans per citation. Figure~\ref{fig:entity_news_cite_dist} shows the
distribution of entities from the different groups. From each sampled
entity, we extract all \emph{citing paragraphs} that contain either a
\emph{web} or \emph{news} citation. Our sample consists of 509 citing
paragraphs from 134 entities.

Furthermore, since a paragraph may have more than one citation, in our
sampled citing paragraphs, we have an average of 4.4 citations per
paragraph, which finally resulted in 408 unique
paragraphs. Table~\ref{tbl:dataset_stats} shows the stats of the
dataset.

\begin{table}
	\centering
	\begin{tabular}{l c c c}
	\toprule
	\emph{type} & \emph{avg. $|s|$} & \emph{avg. $|\delta|$} & \emph{avg `covered'}\\
	\midrule
	\emph{news} & 7.76 & 22.55 &	 0.28\\
	\emph{web} & 8.67 & 	23.07 &	0.30\\
	\bottomrule
	\end{tabular}
	\caption{Dataset statistics for citing paragraphs, distinguishing between \emph{web} and \emph{news} references, showing the average number of sentences, fragments, and covered fragments.}
	\label{tbl:dataset_stats}
\end{table}

\subsection{Ground Truth}\label{subsec:ground_truth}

\paragraph{Setup.}

For the ground truth, the citation span of $c$ in paragraph $p$ was manually determined by labeling
each fragment in $p$ with the binary label \emph{covered} or \emph{not-covered}. 

We set strict guidelines that help us generate reliable ground-truth
annotations. We follow two main guidelines: (i) requirement to read
and comprehend the content in $c$, and (ii) matching of the textual
fragments from $p$ as either being supported \emph{explicitly} or
\emph{implicitly} in $c$.\footnote{We excluded cases  where the citation is not
appropriate for the paragraph at all. This
  is, for example, the case  when the language of $c$ is not English.}

The entire dataset was
carefully annotated by the first author. Later, a second
annotator annotated a 10\% sample of the dataset with an inter-rater
agreement of $\kappa=.84$. We chose not to use crowd-sourcing as
the task is very complex and hard to divide into small independent tasks. 
Since the task requires reading and comprehending the entire content
in $c$ and $p$, it takes on average up to 2.4 minutes to perform the
evaluation for a single item. In future, it would be worthwhile to conduct more large-scale
annotation exercises.

\paragraph{Citation Span Stats.} Following the definition in Equation~\ref{eq:cite_span}, we determine the citation span at the sub-sentence granularity level. Table~\ref{tbl:gt_stats} shows the distribution of citations falling into the specific spans for the citing paragraphs. We note that the majority of citations have a span between half a sentence and up to a sentence, yet, the remainder of more than 20\% of citation span across multiple sentences in such paragraphs. 

We define the citation span as the ratio of sub-sentences which are
covered by a given citation over the total number of sub-sentences in
the sentence, consequentially in the citing paragraph. That is, a citation
is considered to have a span of one sentence if it covers all its
sub-sentences.

\begin{equation}\label{eq:cite_span}
span(c,p) = \sum\limits_{s \in p}\frac{\#\delta^{s} \in \mathcal{S'}}{\#\delta^{s}}	
\end{equation}
where $\delta^s$ represents a sequence in sentence $s\in p$, which are part of the the ground-truth.

\begin{table}[h!]
\centering\small
\begin{tabular}{p{0.5cm} p{0.5cm} l l l l p{0.6cm}}
\toprule
 & total & $\leq .5$ & $(.5,1]$ & $(1,2]$ & $(2,5]$ & $>5$\\
 \midrule

news & 318 & 35 & 201 & 54  & 22 & 6\\
web & 191 & 13 & 121 & 27 & 25 & 6\\
\bottomrule
\end{tabular}
\caption{Citation span distribution based on the number of sub-sentences in the citing paragraph. }
\label{tbl:gt_stats}
\end{table}

\begin{figure}[ht!]
\centering
\includegraphics[width=0.8\columnwidth]{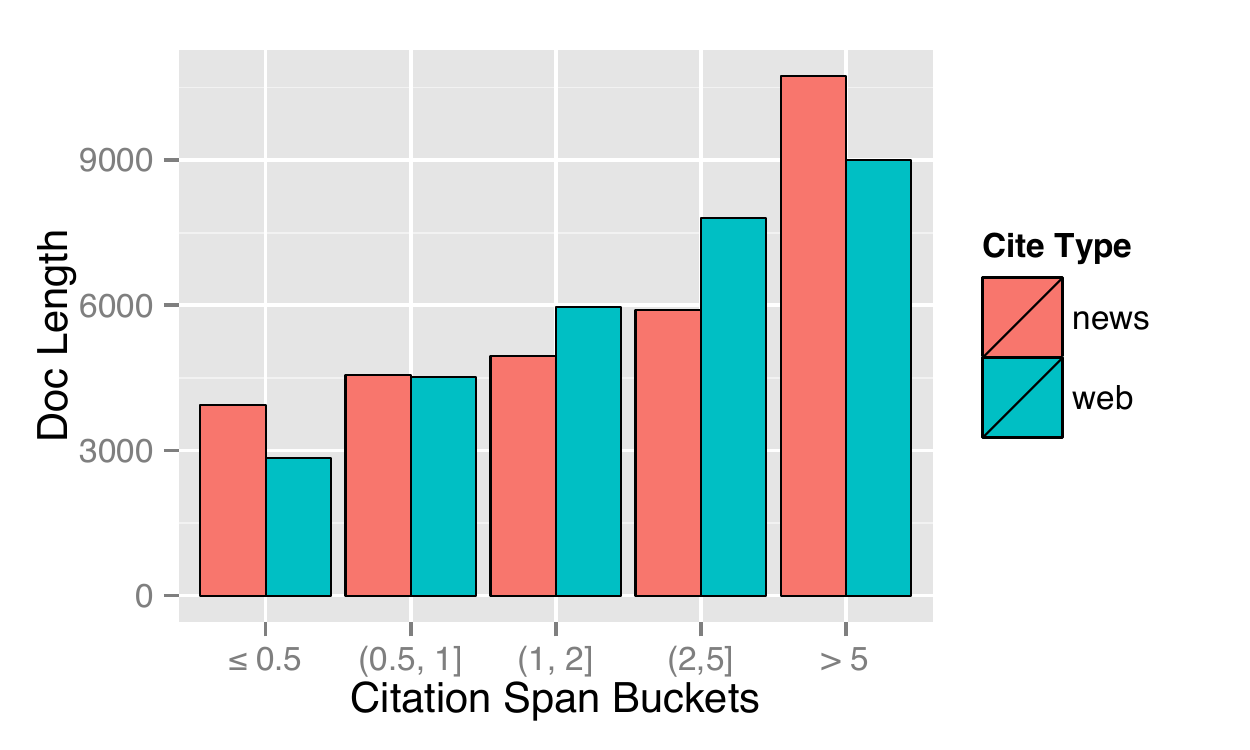}
\caption{Average document length for the different span buckets for citation types \emph{web} and \emph{news}.}
\label{fig:doc_length_cite_span}
\end{figure}

In Figure~\ref{fig:doc_length_cite_span}, we analyze a possible factor
in the variance of the citation span. It is evident that for longer
cited documents the span increases. This is intuitive since such documents
carry more information and consequentially their span in the citing
paragraphs can be larger. An example is the Wikipedia article
\texttt{2008 US Open (tennis)} which has a citing paragraph with a
citation span of 7 sentences for an article of 30k characters
long\footnote{\url{http://news.bbc.co.uk/sport1/hi/tennis/7601195.stm}}. We
encoded this in the \emph{citation} features $f^c$.

Additionally, within the different citation spans we analyze how many
of them contain \emph{skips} for two cases: (i) skip a fragment within
a sentence, and (ii) skip sentences in $p$. The results for both cases
are presented in Table~\ref{tbl:gt_skips}.

\begin{table}[h!]
\centering\small
\begin{tabular}{r l l l l}
\toprule
\emph{span} & \multicolumn{2}{c}{\emph{news}} & \multicolumn{2}{c}{\emph{web}}\\
 & \emph{skip} $\delta$ & \emph{skip} $s$ & \emph{skip} $\delta$ & \emph{skip} $s$\\
 \midrule
$\leq 0.5$ & 6\% & - & - & -\\
$(0.5,1]$ & - & - & - & 1\%\\
$(1,2]$ & - & 8\% & - & 19\%\\
$(2,5]$ & 5\% & 18\% & - & 21\%\\
$>5$ & - & 20\% & - & 67\%\\
\bottomrule
\end{tabular}
\caption{The percentage of citations in a span with \emph{fragment skips} and \emph{sentence skips}.}
\label{tbl:gt_skips}
\end{table}

From the results in Table~\ref{tbl:gt_stats} and \ref{tbl:gt_skips} we
see that simple heuristics on selecting complete sentences or
selecting consecutive sequences do not account for the different
citation span cases and skips at the sentence and paragraph
level. This leads to suboptimal results and introduces erroneous
spans. Furthermore, we find that in 3.7\% of the cases in our
ground-truth, the citation spans include fragments after the citation marker.

\subsection{Baselines}\label{subsec:baselines}

We consider the following baselines as competitors for our citation span approach. 

\textbf{Inter-Citation Text -- IC.} The span consists of sentences
which start either at the beginning of the paragraph or at the end of
a previous citation. The granularity is at the sentence level.

\textbf{Citation-Sentence-Window -- CSW.} The span consists of sentences in a window of +/- 2 sentences from the citing sentence~\cite{DBLP:journals/ipm/OConnor82}. The other sentences are included if they contain specific cue words in fixed positions. 

\textbf{Citing Sentence -- CS.} The span consists of only the \emph{citing sentence}. 

\textbf{Markov Random Fields - MRF.} MRFs~\cite{DBLP:conf/acl/QazvinianR10} model two functions. First, \emph{compatibility}, which measures the similarity of sentences in $p$, and as such allows to extract non-citing sentences. Second, the \emph{potential}, which measures the similarity between sentences in $c$ with sentences in $p$. We use the provided implementation by the authors.

\textbf{Citation Span Plain -- CSPC.} A plain classification setup using the features in Section~\ref{sec:approach}, where the sequences are classified in isolation. We use Random Forests~\cite{DBLP:journals/ml/Breiman01} and evaluate them with 5-fold cross validation.

\subsection{Citation Span Approach Setup -- CSPS}
For our approach $CSPS$ as mentioned in Section~\ref{sec:approach}, we
opt for linear-chain CRFs and use the implementation in
\cite{okazaki2007crfsuite}. We evaluate our models using 5-fold cross
validation, and learn the optimal parameters for the CRF model through
the L-BFGS approach~\cite{liu1989limited}.

\subsection{Evaluation Metrics}\label{subsec:metrics}

We measure the performance of the citation span approaches through the following metrics. We will denote with $W'$ the sampled entities, with $\mathbf{p} =\{p_c, \ldots\}$ ($p_c$ refers to $\langle p, c\rangle$) the set of sampled paragraphs from $e$, and with $|\mathbf{p}|$ the total items from $e$.

\textbf{Mean Average Precision -- $MAP$.} First, we define \emph{precision} for $p_c$ as the ratio $P(p_c) = {|\mathcal{S}'\cap\mathcal{S}^{t}|}/{|\mathcal{S}'|}$ of fragments present in $\mathcal{S}'\cap \mathcal{S}^t$ over $\mathcal{S}'$. We measure MAP as in Equation~\ref{eq:map}. 
\begin{equation}\label{eq:map}\small
MAP = \frac{1}{|W'|}\sum_{e\in W'}\frac{\sum_{p_c \in \mathbf{p}}P(p_c)}{|\mathbf{p}|}
\end{equation}

\textbf{Recall -- $R$.} We measure the recall for $p_c$ as the ratio $\mathcal{S}'\cap\mathcal{S}^t$ over  all fragments in $\mathcal{S}^t$, $R(p_c)=|\mathcal{S}'\cap\mathcal{S}^t|/|\mathcal{S}^t|$. We average the individual recall scores for $e\in W'$ for the corresponding $\mathbf{p}$.
\begin{equation}\label{eq:recall}\small
R = \frac{1}{|W'|}\sum_{e\in W'}\frac{\sum_{p_c \in \mathbf{p}}R(p_c)}{|\mathbf{p}|}
\end{equation}

\textbf{Erroneous Span -- $\Delta$.} We measure the number of extra \emph{words} or extra \emph{sub-sentences} (denoted with $\Delta_w$ and $\Delta_\delta$) added by text fragments that are not part of the ground-truth $\mathcal{S}^t$. The ratio is relative to the number of words or sub-sentences in the ground-truth for $p_c$. We compute $\Delta_w$ and $\Delta_\delta$ in Equation~\ref{eq:extra_words} and~\ref{eq:extra_subs}, respectively.
\begin{equation}\label{eq:extra_words}\small
\Delta_w = \frac{1}{|W'|}\sum_{e\in W'}\frac{1}{|\mathbf{p}|}\sum_{p_c \in \mathbf{p}}\frac{\sum_{\delta\in \mathcal{S}'\setminus \mathcal{S}^t}words(\delta)}{\sum_{\delta\in \mathcal{S}^t}words(\delta)}
\end{equation}

\begin{equation}\label{eq:extra_subs}\small
\Delta_\delta = \frac{1}{|W'|}\sum_{e\in W'}\frac{1}{|\mathbf{p}|}\sum_{p_c \in \mathbf{p}}\frac{|\mathcal{S}'\setminus \mathcal{S}^t|}{|\mathcal{S}^t|}
\end{equation}

\section{Results and Discussion}\label{sec:results}

\subsection{Citation Span Robustness}\label{subsec:robustness}
Table~\ref{tbl:cspan_results} shows the results for the different approaches on determining the citation span for all  span cases shown in Table~\ref{tbl:gt_stats}. 

\paragraph{Accuracy.} Not surprisingly, the baseline approaches perform reasonably well. $CS$ which selects only the citing sentence achieves a reasonable $MAP=0.86$ and similar recall. A slightly different baseline $CSW$ achieves comparable scores with $MAP=0.85$. This is due to the inherent span structure in Wikipedia, where a large portion of citations span up to a sentence (see Table~\ref{tbl:gt_stats}). Therefore, in approximately 64\% of the cases the baselines will select the correct span. For the cases where the span is more than a sentence, the drawback of these baselines is in coverage. We show in the next section a detailed decomposition of the results and highlight why even in the simpler cases, a sentence level granularity has its shortcomings due to sequence skips as shown in Table~\ref{tbl:gt_skips}.

Overall, when comparing $CS$ as the best performing baseline against our approach $CSPS$, we achieve an overall score of $MAP=0.83$ (a
slight decrease of 3.6\%), whereas in term of F1 score, we have a decrease of 9\%. The plain-classification approach $CSPC$ achieves similar  score with $MAP=0.86$, whereas in terms of F1 score, we have a decrease of 8\%.  As described above and as we will see later on in
Table~\ref{tbl:cspan_bucket_results}, the overall good performance of the baseline approaches can be  attributed to the citation span distribution in our ground-truth.

On the other hand, an interesting observation is that sophisticated approaches, geared towards scientific domains like $MRF$
perform poorly. We attribute this to  \emph{language style}, i.e., in Wikipedia there are no explicit citation hooks that are present in
scientific articles. Comparing to $CSPS$, we outperform $MRF$ by a large margin with an increase in $MAP$ by $84\%$.

When comparing the sequence classifier $CSPS$ to the plain classifier $CSPC$, we see a marginal difference of 1.3\% for \emph{F1}. However, it will become more evident later that classifying jointly the text fragments for the different span buckets, outperforms the plain classification model.

\begin{table}[h!]
\centering\small
\begin{tabular}{r l l l r r}
\toprule
  & \multicolumn{1}{c}{MAP} & \multicolumn{1}{c}{R} & \multicolumn{1}{c}{F1} & \multicolumn{1}{c}{$\Delta_{w}$} & \multicolumn{1}{c}{$\Delta_{\delta}$}\\
 \midrule
  {MRF}  & 0.45 & 0.78 & 0.56 &  308\% & 278\%\\
  {IC}   & 0.72 & \textbf{0.94} & 0.77 & 113\% & 115\%\\
  {CSW}  & 0.85 & 0.84 & \textbf{0.82} & 38\% & 31\%\\
  {CS}   & \textbf{0.86} & 0.84 & \textbf{0.82} & 35\% & 27\%\\
  {CSPC} & \textbf{0.86} & 0.68 & 0.76 &  \textbf{26}\% & \textbf{23}\%\\
  {CSPS} & 0.83 & 0.69 & 0.75 &  32\% & 24\%\\

 \bottomrule
\end{tabular}
\caption{Evaluation results for the different citation span approaches.}
\label{tbl:cspan_results}
\end{table}

\paragraph{Erroneous Span.} One of the major drawbacks of competing approaches is the granularity at which the span is determined. This leads to erroneous spans. From Table~\ref{tbl:gt_stats} we see that approximately in $\sim$10\% of the cases the span is at sub-sentence level, and in 28\% the span is more than a sentence. 

The best performing baseline $CS$ has an erroneous span of $\Delta_w=35\%$ and $\Delta_\delta=27\%$, in terms of extra words and
sub-sentences, respectively. That is, nearly half of the determined span is erroneous, or in other words it is not covered in the provided
citation. The $MRF$ approach due to its poor $MAP$ score provides the largest erroneous spans with $\Delta_w=308\%$ and $\Delta_\delta=278\%$. The amount of erroneous span is unevenly distributed, that is, in cases where the span is not at the sentence level granularity the amount of erroneous span increases. A detailed analysis is provided in the next section.

Contrary to the baselines, for $CSPS$ and similarly for $CSPC$, we achieve the lowest erroneous spans with $\Delta_w=32\%$ and
$\Delta_\delta=26\%$, and $\Delta_w=24\%$ and $\Delta_w=23\%$, respectively.

Compared to the remaining baselines, we achieve an overall relative decrease of $9\%$ for $\Delta_w(CSPS)$, and $34\%$ for $\Delta_w(CSPC)$, when compared to the best performing baseline $CS$.

From the \emph{skips} in sequences in Table~\ref{tbl:gt_skips} and the unsuitability of sentence granularity for citation spans, we analyze
the locality of erroneous spans w.r.t to the sequence that contains $c$, specifically the distribution of erroneous spans \emph{preceding}
and \emph{succeeding} it. For the $CS$ baseline, $71\%$ of the total erroneous spans are added by sequences preceding the citing sequence,
contrary to $35\%$ which succeed it. In the case of $CSPS$, we have only $9\%$ of erroneous spans (for $\Delta_\delta$) preceding the citation.

\subsection{Citation Span and Feature Analysis}\label{subsec:cite_struct_feature_ablation}

We now analyze how the approaches perform for the different citation spans in Table~\ref{tbl:gt_stats}\footnote{The models were retrained and tested for the different buckets with 5-fold cross validation.}. Additionally, we analyze how our approach $CSPS$ performs when determining the span without access to the content of $c$.

\begin{table*}[ht!]
	\centering\small
	\begin{tabular}{l l l l l l l l l l l l l}
		\toprule
		& \multicolumn{3}{c}{$\leq 0.5$} & \multicolumn{3}{c}{$(0.5,1]$} & \multicolumn{3}{c}{$(1,2]$} & \multicolumn{3}{c}{$> 2$}	\\[1.5ex]
		& MAP & R & F1 & MAP & R & F1  & MAP & R & F1 & MAP & R & F1 \\
		\midrule
		MRF 	& 0.15 & 0.88 & 0.27 & 0.44 & 0.80 & 0.61 & 0.59 & 0.74 & 0.57 & 0.59 & 0.63 & 0.55\\
		IC 	    & 0.32 & \textbf{1.00} & 0.45 & 0.77 & \textbf{0.99} & 0.83 & 0.73 & \textbf{0.84} & 0.74 & 0.72 & \textbf{0.81} & \textbf{0.73}\\
		CSW 	& 0.38 & \textbf{1.00} & 0.54 & 0.93 & 0.98 & 0.96 & 0.88 & 0.54 & 0.65 & 0.79 & 0.34 & 0.43\\
		CS 	    & 0.40 & \textbf{1.00} & 0.56 & 0.94 & 0.98 & 0.97 & 0.90 & 0.53 & 0.65 & \textbf{0.80} & 0.32 & 0.42 \\[1.5ex]
		
		\midrule 
		
		CSPC 	& 0.85 & 0.53 & 0.65 & \textbf{0.96} & 0.97 & 0.97 & \textbf{0.96} & 0.68 & 0.79 & 0.71 & 0.65 & 0.68\\
		CSPS 	& \textbf{0.87**} & 0.56 & \textbf{0.68**} & \textbf{0.96} & 0.98 & \textbf{0.98} & 0.88 & 0.73 & \textbf{0.80*} & 0.74 & 0.72 & 0.70 \\[0.8ex]

		\midrule
		$\Delta_{F1}$ CSPS & 
		 & & $\blacktriangle 21\%$ & 
		 & & $0\%$ & 
		 & & $\blacktriangle 8\%$ & 
		 & & $\blacktriangledown 4\%$\\

		\bottomrule
	\end{tabular}
	\caption{Evaluation results for the citation span approaches for the different span cases. For the results of $CSPS$ we compute the relative increase/decrease of $F1$ score compared to the best result (based on $F1$) from the competitors. We mark in bold the best results for the evaluation metrics, and indicate with ** and * the results which are highly significant ($p<0.001$) and significant ($p<0.05$) based on \emph{t-test} statistics when compared to the best performing baselines (CS, IC, CSW, MRF) based on F1 score, respectively.}
	\label{tbl:cspan_bucket_results}
\end{table*}

\paragraph{Citation Span.} Table~\ref{tbl:cspan_bucket_results} shows the results for the approaches under comparison for all the citation span cases. In the case where the citation spans up to a sentence, that is $(0.5, 1]$, which presents the simplest citation span case, the baselines perform reasonably well. This is due to the heuristics they apply to determine the span, which in all cases includes the \emph{citing sentence}. In terms of $F1$ score, the baseline $CS$ achieves a highly competitive score of $F1=0.97$. Our approach $CSPS$ in this case has slight increase of 1\% for $F1$ and an increase of 3\% for $MAP$. $CSPC$ achieves a similar performance in this case.

However, for the cases where the span is at the sub-sentence level or
across multiple sentences, the performance of baselines drops
drastically. In the first bucket ($\leq 0.5$) which accounts for 9\%
of ground-truth data, we achieve the highest score with $MAP=0.87$,
though with lower recall than the competitors with $R=0.56$. The
reason for this is that the baselines take complete sentences, thus,
having perfect recall at the cost of accuracy. In terms of $F1$ score
we achieve 21\% better results than the best performing baseline  $CS$.

For the span of $(1,2]$ we maintain an overall high accuracy and recall, and have the highest $F1$ score. The improvement is 8\% in terms of $F1$ score.
Finally, for the last case where the span is more than 2 sentences, we achieve  $MAP=0.74$, a marginal increase of 3\%, however with lower recall, which results in an overall decrease of 4\% for $F1$. The statistical significance tests are indicated with ** and * in Table~\ref{tbl:cspan_bucket_results}.

\begin{figure}[h]
	\includegraphics[width=1.0\columnwidth]{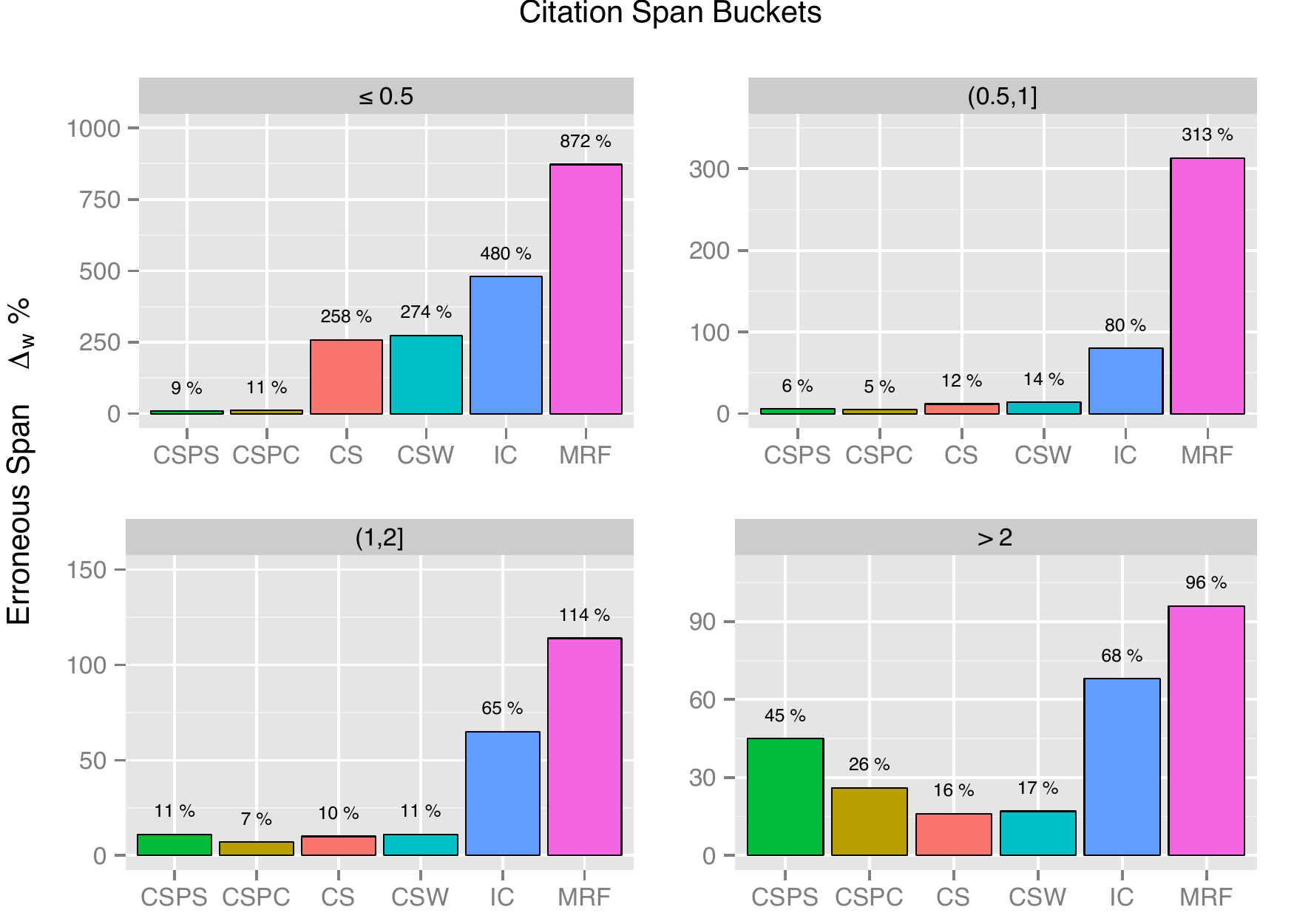}
	\caption{Erroneous spans for the different citation span buckets. The y-axis presents the $\Delta_w$ whereas in the  x-axis are shown the different approaches.}
	\label{fig:spanbucket_results}
\end{figure}

\paragraph{Erroneous Span.} Figure~\ref{fig:spanbucket_results} shows the erroneous spans in terms of words for the metric $\Delta_w$ for all citation span cases. It is noteworthy that the amount of error can be well beyond 100\% due to the ratio of the suggested span and the actual span in our ground-truth, which can be higher.

In the first bucket (span of $\leq 0.5$) with granularity less than a sentence, all the competing approaches introduce large erroneous
spans. For $CSPS$ we have a $MAP=0.87$, and consequentially we have the lowest $\Delta_w=9\%$, while for $CSPC$ we have only $\Delta_w=11\%$. In contrast, the non-ML competitors introduce a minimum of $\Delta_w(CS)=182\%$, with MRFs having the highest error.  We also perform well in the bucket $(0.5,1]$. For larger spans, for instance, for $(1,2]$, we are still slightly better, with roughly 3\% less erroneous span when comparing $CSPC$ and $CS$. However, only in the case of spans with $>2$, we perform below the CS baseline. Despite, the smaller erroneous span, the $CS$ baseline never includes more than one sentence, and as such it does not include many erroneous spans for the larger buckets. However, it is by definition unable to recognize any longer spans.

\paragraph{Feature Analysis.} It is worthwhile to investigate the performance gains in determining the citation span without analyzing the content of the citation. The reason for this is that there are several citation categories for which access to the source cannot be easily automated. Models which can determine the span accurately without the actual content have the advantage of generalizing to other citation sources (e.g. \emph{books}) for which the evaluation is more challenging.\footnote{At worst, one needs to read and comprehend
  the entire book to determine if a fragment is covered by the citation.}

Here, we disregard the citation features from
Section~\ref{subsec:cite_features}. In terms of $MAP$, we have a
slight decrease with $MAP=0.82$ when compared to the model with the
citation features. For recall we have a drop of 3\%, resulting in
$R=0.67$. 

This shows that by solely relying on the structure of the citing paragraph and other structural and discourse features we can perform the task with reasonable accuracy.

\section{Conclusion}\label{sec:conclusion}

In this work, we tackled the problem of determining the fine-grained
citation span of references in Wikipedia. We started from the
\emph{citing paragraph} and decomposed it into sequences consisting of
sub-sentences. To accurately determine  the span we proposed features
that leverage the structure of the paragraph, discourse and temporal
features, and finally analyzed the similarity between the citing
paragraph and the citation content.

We introduce both a standard classifier as well as a sequence classifier using
a linear-chain CRF model.
For evaluation we manually annotated a ground-truth dataset of
509 citing paragraphs. We reported standard evaluation metrics and
also introduced metrics that measure the amount of erroneous span.

We achieved a $MAP=0.86$, in the case of the plain classification model $CSPC$, and with a marginal difference for $CSPS$ with $MAP=0.83$, across all cases with an  erroneous span of $\Delta_w=26\%$ or $\Delta_w=32\%$, depending on the
model. Thus, we provide accurate means on determining the span and at
the same time decrease the erroneous span by 34\% compared to the best
performing baselines. Moreover, we excel at determining citation spans
at the sub-sentence level.

In conclusion, this presents an initial attempt on solving the
citation span for references in Wikipedia. As future work we foresee a
larger ground-truth and more robust approaches which take into account
factors such as a reference being irrelevant to a citing paragraph and
cases where the evidence for a paragraph is implied rather than
explicitly stated in the reference.

\section*{Acknowledgments} This work is funded by the ERC Advanced Grant ALEXANDRIA (grant no. 339233), and H2020 AFEL project (grant no. 687916).

\end{document}